\documentclass[twocolumn]{article}
\usepackage[T1]{fontenc}
\usepackage{authblk}
\usepackage{geometry}
\usepackage[misc]{ifsym} 
\usepackage{amsmath}
\title{Automatic assembly of aero engine low pressure tur-bine shaft based on 3D vision measurement}

\author{Jiaxiang Wang,$^{1\ast}$ Kunyong Chen,$^{1}$ \\
\normalsize{$^{1}$School of mechanical and power engineering, Shanghai Jiaotong University}\\
\normalsize{800 Dongchuan Road, Minhang District, Shanghai， 200240,  china}\\
\normalsize{$^\ast$ E-mail:  jiaxiang\_wang@sjtu.edu.cn}
}

\date{} %
\usepackage{graphicx}

\begin{document}
\twocolumn[
\begin{@twocolumnfalse}
\maketitle
\section*{Abstract}
 In order to solve the problem of low automation of Aero-engine Turbine shaft assembly and the difficulty of non-contact high-precision measurement, a structured light binocular measurement technology for key components of aero-engine is proposed in this paper. Combined with three-dimensional point cloud data processing and assembly position matching algorithm, the high-precision measurement of shaft hole assembly posture in the process of turbine shaft docking is realized. Firstly, the screw thread curve on the bolt surface is segmented based on PCA projection and edge point cloud clustering, and Hough transform is used to model fit the three-dimensional thread curve. Then the preprocessed two-dimensional convex hull is constructed to segment the key hole location features, and the mounting surface and hole location obtained by segmentation are fitted based on RANSAC method. Finally, the geometric feature matching is used the evaluation index of turbine shaft assembly is established to optimize the pose. The final measurement accuracy of mounting surface matching is less than 0.05mm, and the measurement accuracy of mounting hole matching based on minimum ance optimization is less than 0.1 degree. The measurement algorithm is implemented on the automatic assembly test-bed of a certain type of aero-engine low-pressure turbine rotor. In the narrow installation space, the assembly process of the turbine shaft assembly, such as the automatic alignment and docking of the shaft hole, the automatic heating and temperature measurement of the installation seam, and the automatic tightening of the two guns, are realized in the narrow installation space Guidance, real-time inspection and assembly result evaluation.\\
\end{@twocolumnfalse}]
\section{Introduction}
In the complex structure of aero-engine, there are a lot of assembly joint surfaces with bolt fastening connection, such as engine casing, rotor disc, turbine shaft, etc It is easy to bump or even block in the process of matching. At the same time, the low-pressure turbine shaft surface coating can not install the target, and the field light source environment is more complex. Based on the above considerations, the measurement scheme based on structured light binocular is adopted, and the current assembly pose is calculated based on 3D point cloud data. 

Firstly, segmentation and model fitting of spatial 3D thread curve is primary and important. In the research of free-form curve and surface modeling, Andr\cite{1} used the method of resampling and pre fairing to study the spline approximation of spatial curve. Zhu\cite{2} and others used curvature and torsion to detect the feature points of the contour, and minimized the matching error function to realize the curve matching. Secondly, key feature extraction and segmentation. S. belongie proposed a 3D feature local descriptor, which is robust to noise and clutter. Finally, the matching evaluation and Optimization Based on the geometric characteristics of the point cloud. Huang et al. Used the point cloud information measured by the laser tracker in the docking assembly of aircraft large parts to establish the evaluation index of aircraft large parts docking matching. Based on the comprehensive evaluation index, the input value of attitude control system was optimized to improve the docking success rate.
\begin{figure}[ht]
\centering
\includegraphics[scale=0.27]{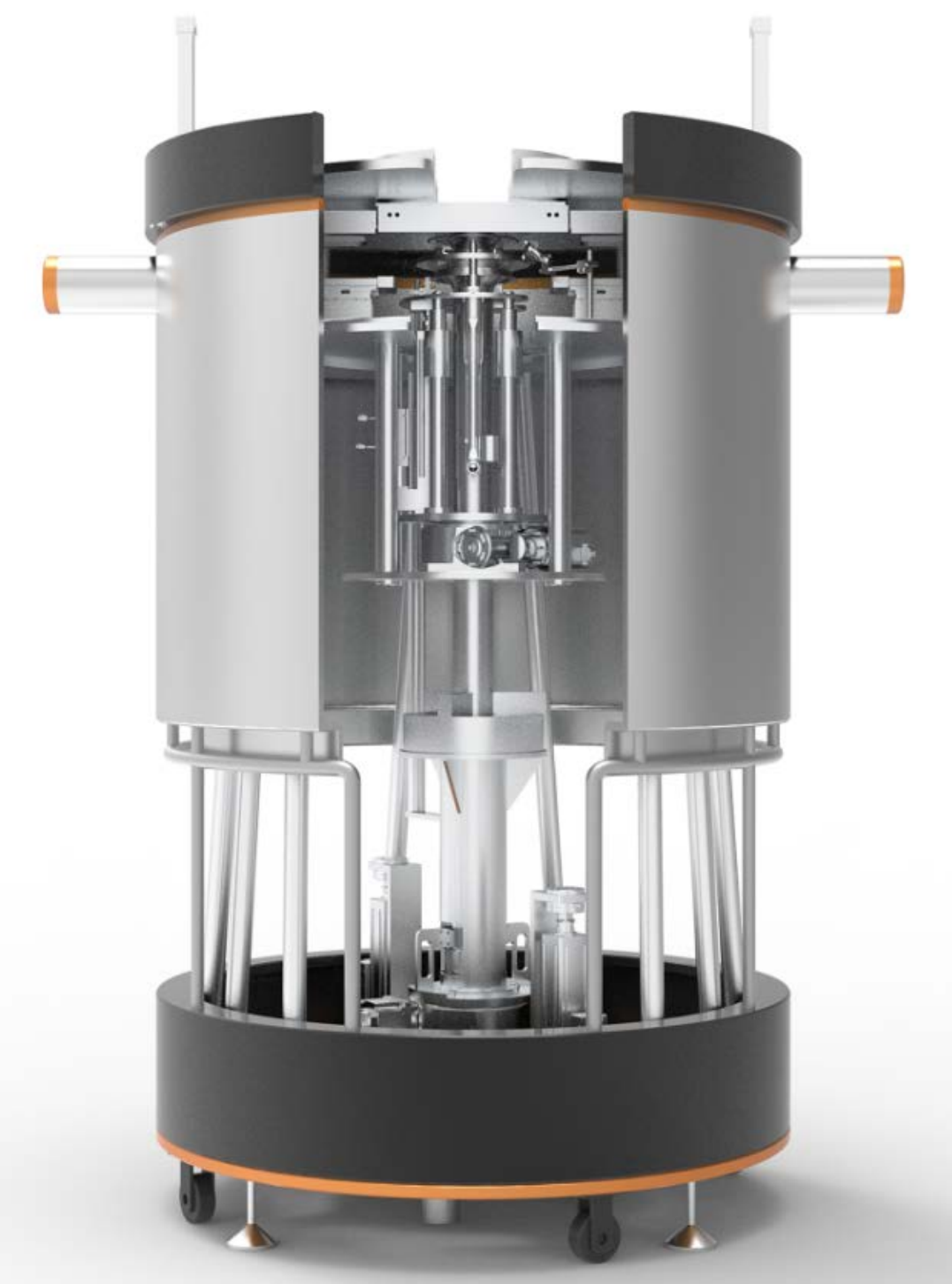}
\caption{Automatic assembly equipment for Aeroengine low pressure turbine shaft}
\label{fig.2}
\end{figure}

\section{Extraction and fitting of 3D thread curve}
In automatic assembly, the recognition and location of thread curve is a difficult problem. It is difficult to obtain accurate depth information based on bolt image.So we use binocular structured light camera to capture the point clouds. 
\begin{figure}[ht]
\centering
\includegraphics[scale=0.25]{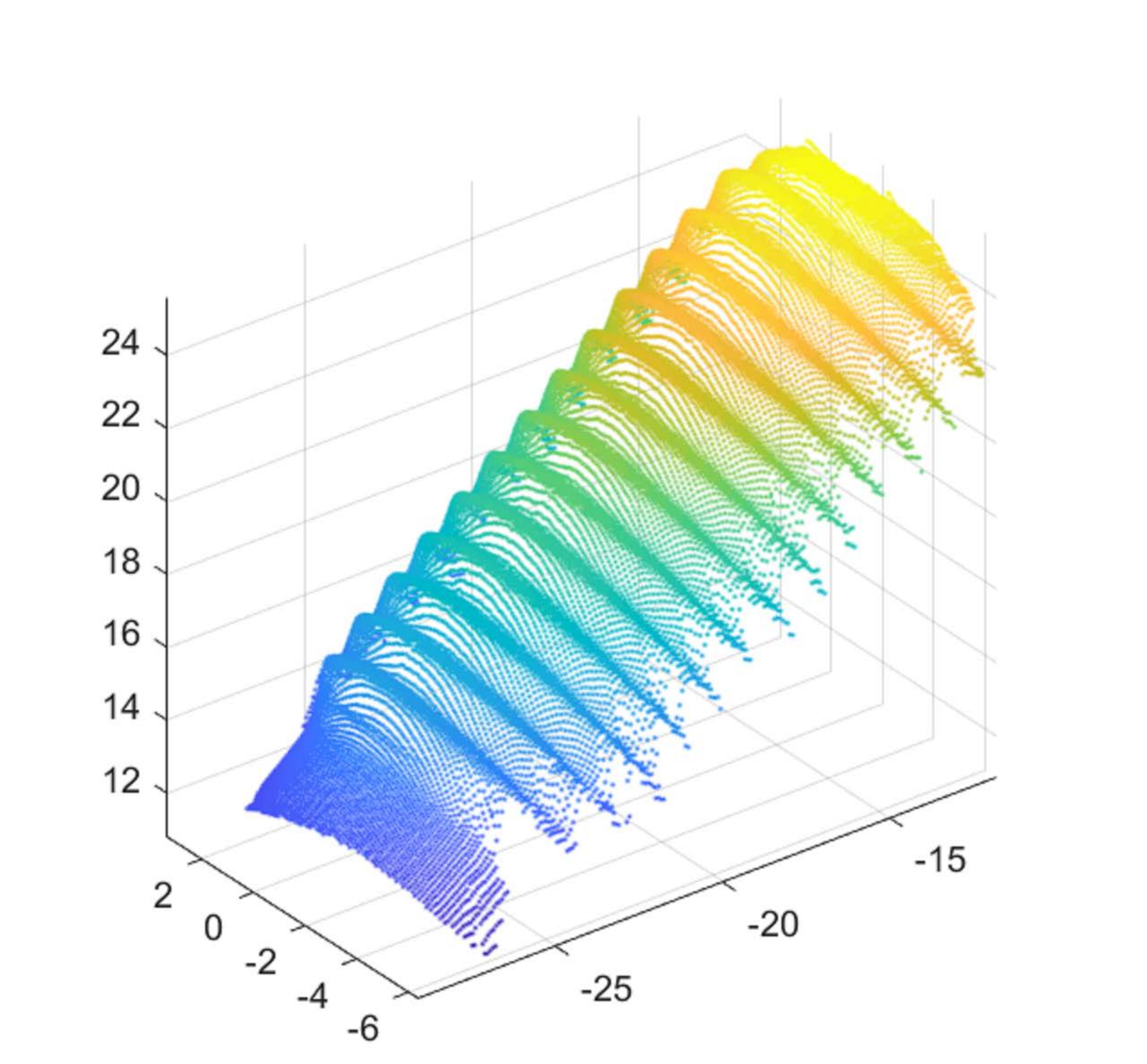}
\caption{Screw point clouds on bolt surface}
\label{fig.3}
\end{figure}

The three-dimensional thread curve of bolt surface has different characteristics from the general space curve, for example, the projection in each principal component space has good geometric characteristics, so PCA can be used to obtain the key thread curve features.\cite{3}Firstly, the point cloud in camera coordinate system is transformed into turbine axis coordinate system to facilitate subsequent processing;

\begin{equation}
\label{eq.1}
{X_T}{\rm{ = }}{X_{{T_{\rm{0}}}}}{\rm{ - }}{T_{{T_{\rm{0}}}}}
\end{equation}
Surface noise can be eliminated by SOR pretreatment,and Select K-Neighborhood of point clouds,Then calculate the parameters of normal distribution:
\begin{equation}
\label{eq.2}
\mu  = \frac{1}{{mk}}\sum\limits_{i = 1}^m {\sum\limits_{j = 1}^k {{d_{ij}}} } ,
\sigma  = \sqrt {\frac{1}{{mk}}\sum\limits_{i = 1}^m {\sum\limits_{j = 1}^k {{{\left( {{d_{ij}} - \mu } \right)}^2}} } } 
\end{equation}
The points outside the $3\sigma$ confidence region are eliminated,
\begin{equation}
\label{eq.3}
\sum\limits_{j = 1}^k {{d_{ij}}}  > \mu  + 3\sigma   , or   \sum\limits_{j = 1}^k {{d_{ij}}}  < \mu  - 3\sigma 
\end{equation}
PCA analysis was performed,
\begin{equation}
\label{eq.4}
\tilde X = \left[ {{{\tilde x}_1}, \cdots ,{{\tilde x}_m}} \right],{\tilde x_i} = {x_i} - \bar x,i = 1, \cdots ,m
\end{equation}

\begin{equation}
\label{eq.5}
H = \tilde X{\tilde X^T} = {U_r}{\Sigma ^2}U_r^T
\end{equation}

\begin{equation}
\label{eq.6}
{U_{\rm{r}}}{\rm{ = }}\left[ {{u_1}{\rm{ }}{u_2}{\rm{ }}{u_3}...{u_r}} \right]
\end{equation}
The main axis is visualized, three main directions are selected for projection, and the projection diagram is obtained

\begin{equation}
\label{eq.7}
{\widetilde x_{23}} = \left[ \begin{array}{l}
u_2^T\\
u_3^T
\end{array} \right]{x_i}
\end{equation}

\begin{equation}
\label{eq.8}
{\widetilde x_{13}} = \left[ \begin{array}{l}
u_1^T\\
u_3^T
\end{array} \right]{x_i}
\end{equation}

\begin{equation}
\label{eq.9}
{\widetilde x_{12}} = \left[ \begin{array}{l}
u_1^T\\
u_2^T
\end{array} \right]{x_i}
\end{equation}
Among them, projection is made in U3 direction, and DBSCAN clustering segmentation is carried out for given starting point of outer circle thread point cloud performed，give the start point

\begin{figure}[ht]
\centering
\includegraphics[scale=0.6]{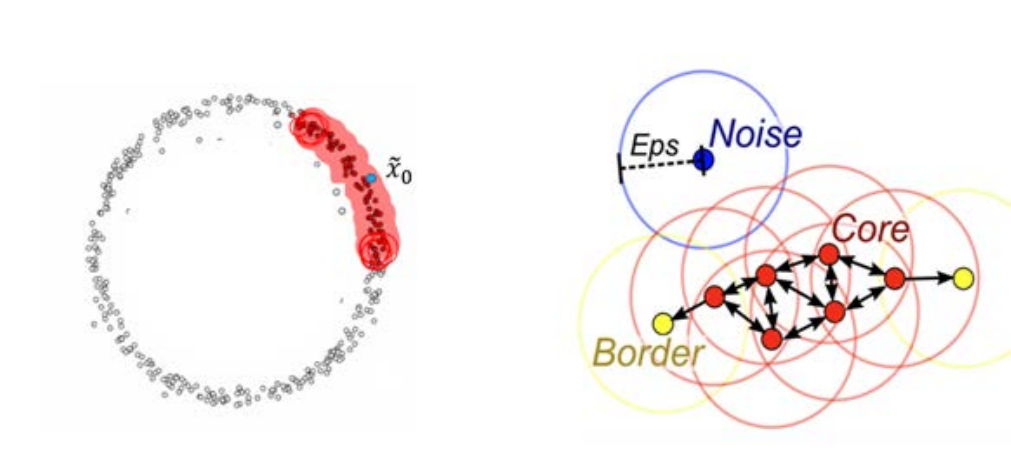}
\caption{DBSCAN clustering segmentation with given starting point}
\label{fig.4}
\end{figure}

\begin{equation}
\label{eq.10}
{\widetilde x_0} = \arg \max \left\| {\widetilde x} \right\|
\end{equation}
set
\begin{equation}
\label{eq.11}
[MinPts] = 4\
\end{equation}
DBSCAN clustering segmentation,and we segment the 3D thread curve.
Next, the model fitting is carried out for the 3D thread curve obtained by segmentation,According to the thread curve equation:
\begin{equation}
\label{eq.12}
x = R\cos \theta, \
y = R\sin \theta ,\
z = \frac{{d\theta }}{{2\pi }}\
\end{equation}{}
Hough change
\begin{equation}
\label{eq.13}
d = \frac{{2\pi z}}{{\arctan \frac{y}{x}}},\
\theta  = \arctan \frac{y}{x},\
R = x\sqrt {{{\left( {\frac{y}{x}} \right)}^2} + 1} \
\end{equation}{}

\begin{figure}[ht]
\centering
\includegraphics[scale=0.36]{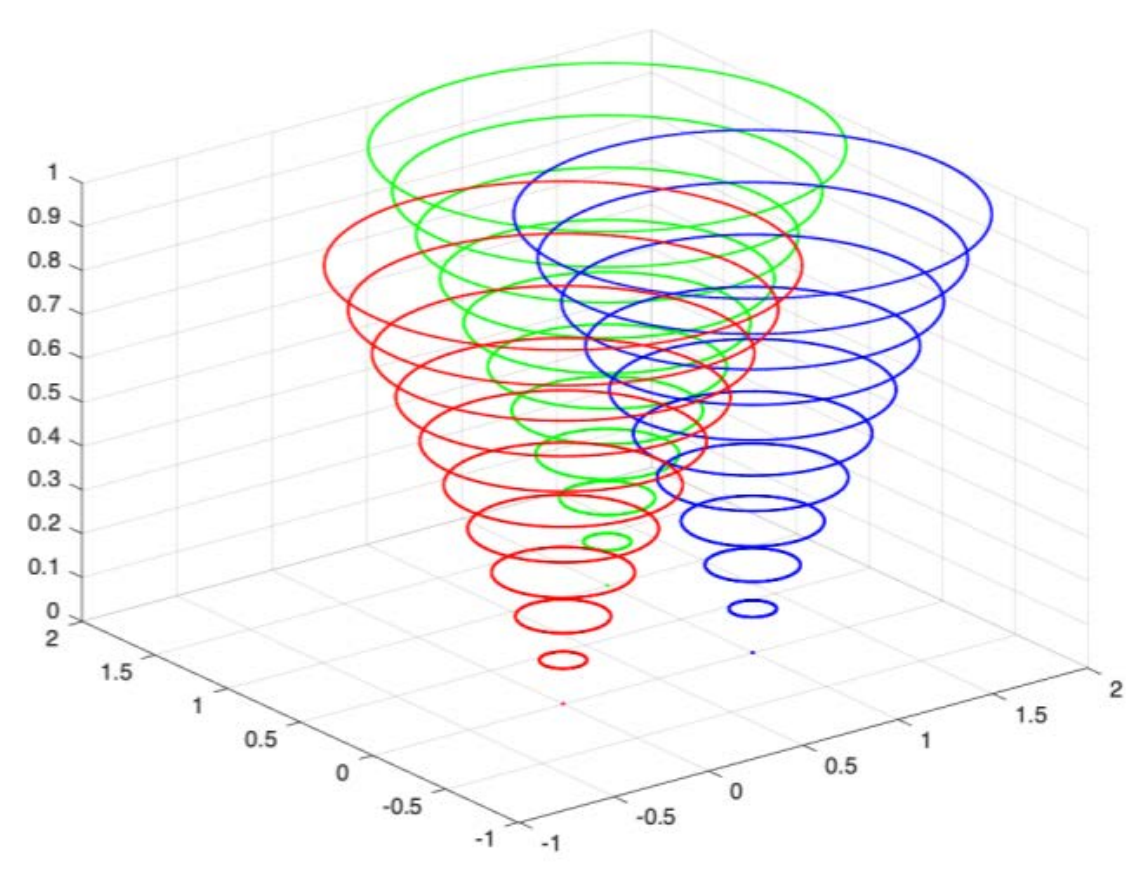}
\caption{Hough space}
\label{fig.5}
\end{figure}
The Hough space of the model parameters of each point cloud coordinate value is brought in.

Set the spatial resolution to 0.01 * 0.01 * 0.01, and search the overlapping area of screw curve parameters in Hough space.\cite{4}For a certain type of aero-engine low-pressure turbine shaft non-standard bolt, the measured thread parameters are as follows:

The triangular thread curve of multi specification bolt with a grade of thread accuracy is extracted and the relevant parameters are settled. The results are as follows:

\section{Assembly key feature extraction and matching}
The feature segmentation and extraction of butt joint surface is based on the point cloud of assembly position after noise reduction for RANSAC cycle segmentation and fitting

a) Randomly select three points
\begin{equation}
\label{eq.14}
{P_i} = ({x_i},{y_i},{z_i}),i = 1,2,3\
\end{equation}{}

b) Determine the plane model
\begin{equation}
\label{eq.15}
Ax + By + Cz + D = 0\
\end{equation}{}
\begin{equation}
\label{eq.16}
{d_i} = \frac{{{n^T}\left( {{p_i} - {p_0}} \right)}}{{\|n{_2}\|}}\
\end{equation}{}

c) Given the threshold to determine the interior point of the mode:
\begin{equation}
\label{eq.17}
{d_i} < {\tau _i}\
\end{equation}{}

d) Repeat a-c steps ${k_1}$, select the plane model with the most interior points

e) Repeat steps a~d for K times in the remaining point cloud, and then $l$ numbers plane models in the point cloud can be obtained，chose:

\begin{equation}
\label{eq.18}
{k_1} = 1000,{k_2} = 1500,{\tau _2} = 0.05,l = 2
\end{equation}{}

\begin{figure}[ht]
\centering
\includegraphics[scale=0.3]{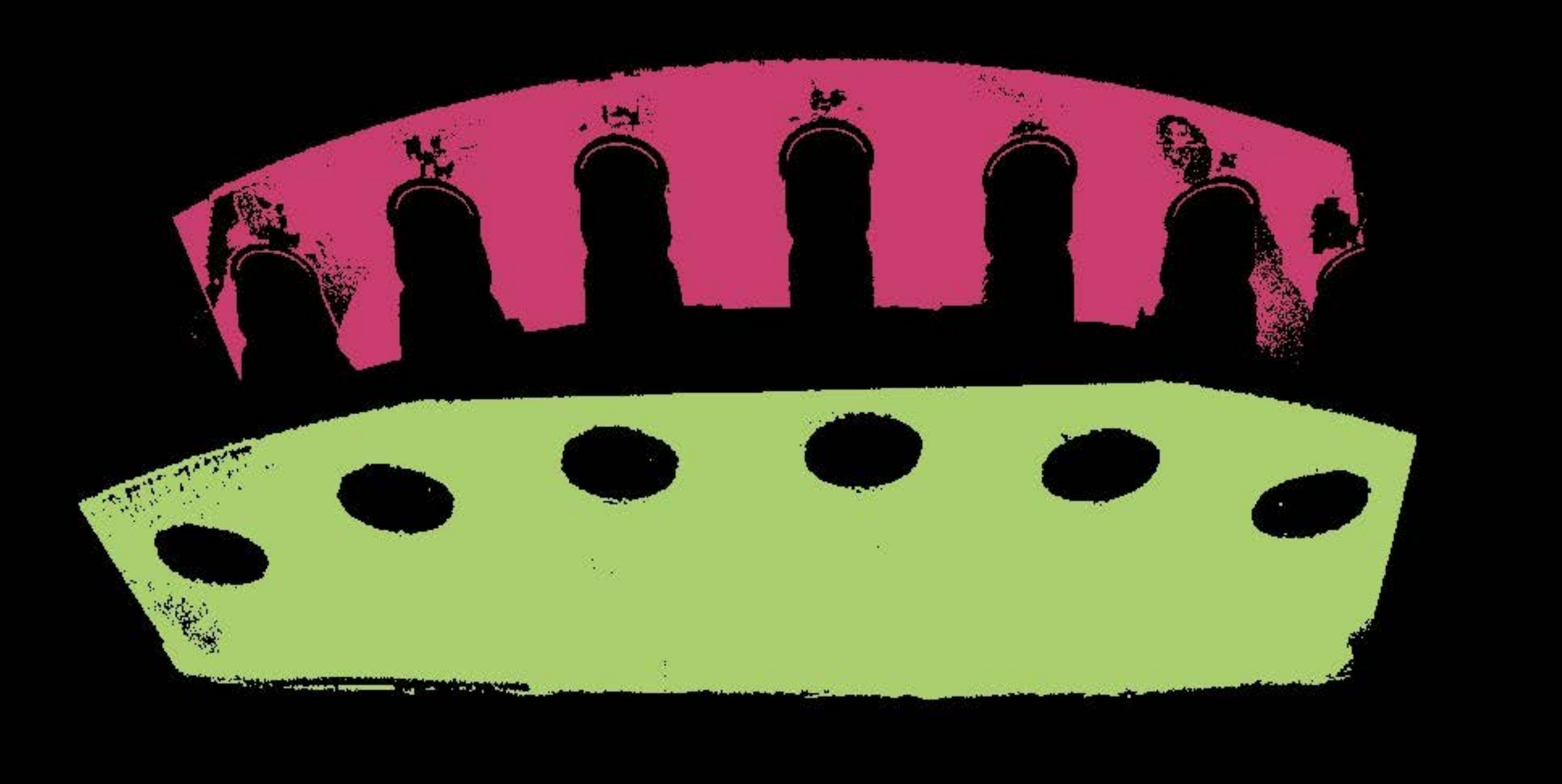}
\caption{Assembly plane point cloud segmentation}
\label{fig.7}
\end{figure}

Firstly, the whole assembly position point cloud is preprocessed, and the installation hole location is pre-searched for the turbine shaft end face point cloud. According to the central axis coordinate value of the hole location obtained by the pre-search, each coordinate value of the convex hull network is determined, and multiple screw hole point clouds are divided.RANSAC cycle segmentation fitting was performed\cite{5}:
\begin{equation}
\label{eq.19}
l = 6,{k_i} = 1000,\tau _i = 0.05,i = 1,2,3,4,5,6
\end{equation}{}
Thus, the axis equation of each hole can be obtained
\begin{equation}
\label{eq.20}
\frac{{x - {x_0}}}{{{n_x}}} = \frac{{y - {y_0}}}{{{n_y}}} = \frac{{z - {z_0}}}{{{n_z}}}
\end{equation}{}
The first is the butt matching of the assembly surface；
\begin{figure}[ht]
\centering
\includegraphics[scale=0.36]{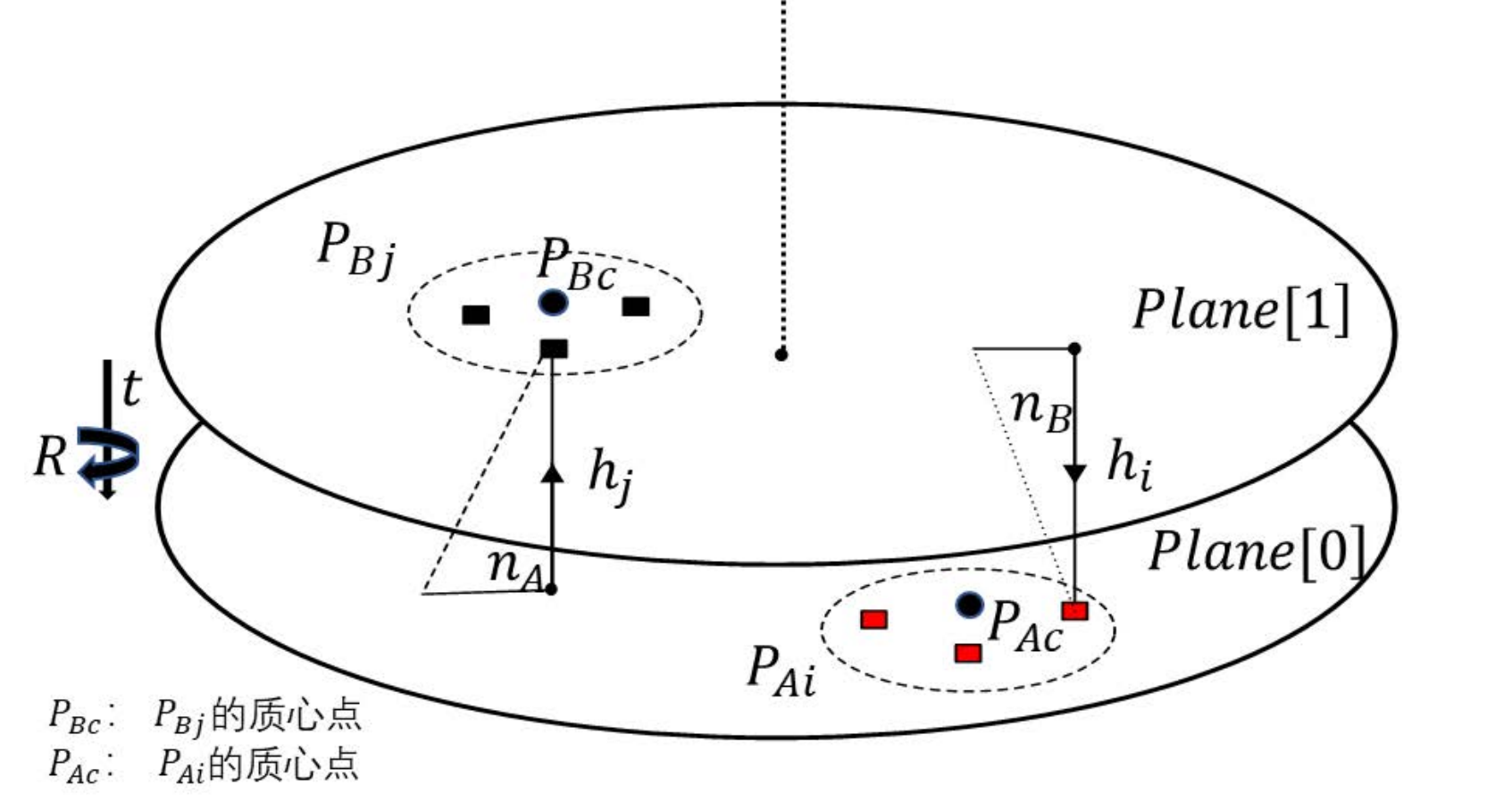}
\caption{Assembly face to face matching}
\label{fig.8}
\end{figure}

Directional distance:
\begin{equation}
\label{eq.21}
{h_i} = n_B^T(R{p_{{A_i}}} + t - {p_{Bc}})
\end{equation}{}
\begin{equation}
\label{eq.22}
{h_j} = {\bf{n}}_A^{\rm{T}}{{\bf{R}}^{\rm{T}}}\left( {{{\bf{p}}_{{B_i}}} - {\bf{R}}{{\bf{p}}_{Ac}} - {\bf{t}}} \right)
\end{equation}{}
The evaluation index of relative deviation of assembly face to face was established
\begin{equation}
\label{eq.23}
{\varepsilon _{pla}}(R,t) = \frac{{{h_{\max }} - \bar h}}{{{h_0}}}
\end{equation}{}
Calculate the rotation displacement of the assembly surface
\begin{equation}
\label{eq.24}
(R,t) = \arg \min {\varepsilon _{pla1}}(R,t)
\end{equation}{}
Secondly, the matching of hole position is carried out, and the bolt surface thread and installation hole position are analyzed\cite{6}
\begin{figure}[ht]
\centering
\includegraphics[scale=0.3]{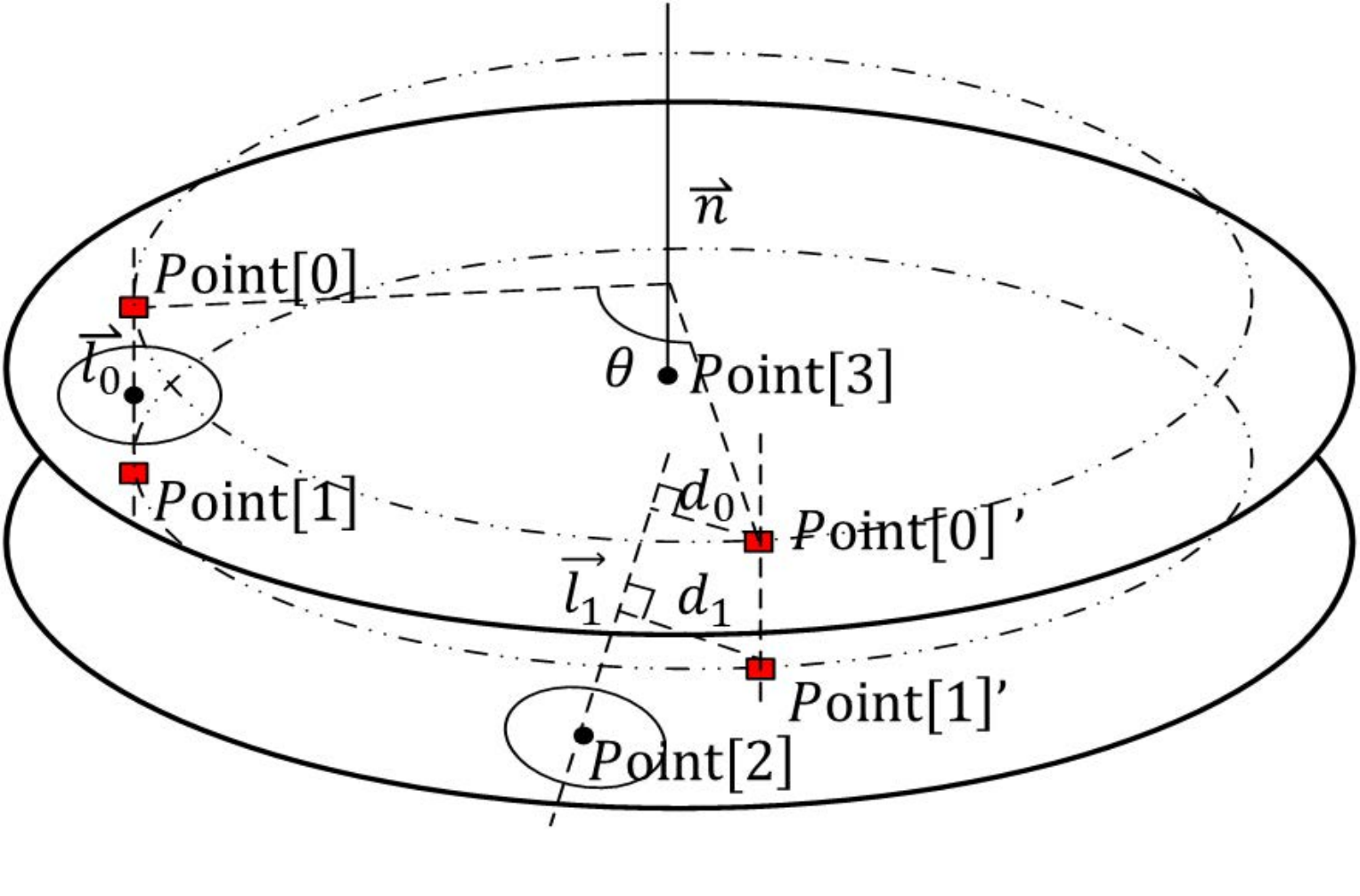}
\caption{Hole position butt matching}
\label{fig.9}
\end{figure}
Coordinates after rotation of bolt central axis:
\begin{equation}
\label{eq.25}
{\widetilde p_k} = {p_O} + ({p_k} - {p_O}){R_n}(\theta )
\end{equation}{}
${n}$ and ${p_0}$ are obtained by cylindrical filtering on the end face of turbine shaft.
Define the relative deviation distance between the screw hole and the center axis of the stud:
\begin{equation}
\label{eq.26}
d(\theta ) = \frac{{{{\overrightarrow n }_l} \times ({{\widetilde p}_k} - {p_l})}}{{{{\overrightarrow n }_l}}}
\end{equation}{}
Evaluation index of relative deviation of hole position butt assembly:
\begin{equation}
\label{eq.27}
{\varepsilon _{cyc}}(\theta ) = \frac{{{d_{\max }} - \overline d }}{{{d_0}}}
\end{equation}{}
Calculate the docking rotation of hole position:
\begin{equation}
\label{eq.28}
\theta  = \arg \min {\varepsilon _{cyc}}(\theta )
\end{equation}{}
So far, the key input$(R,t,\theta )$ is calculated.
\section{System construction and experiment}
According to the requirements of aero-engine assembly process, the aero-engine low-pressure assembly bench was built
\begin{figure}[ht]
\centering
\includegraphics[scale=0.27]{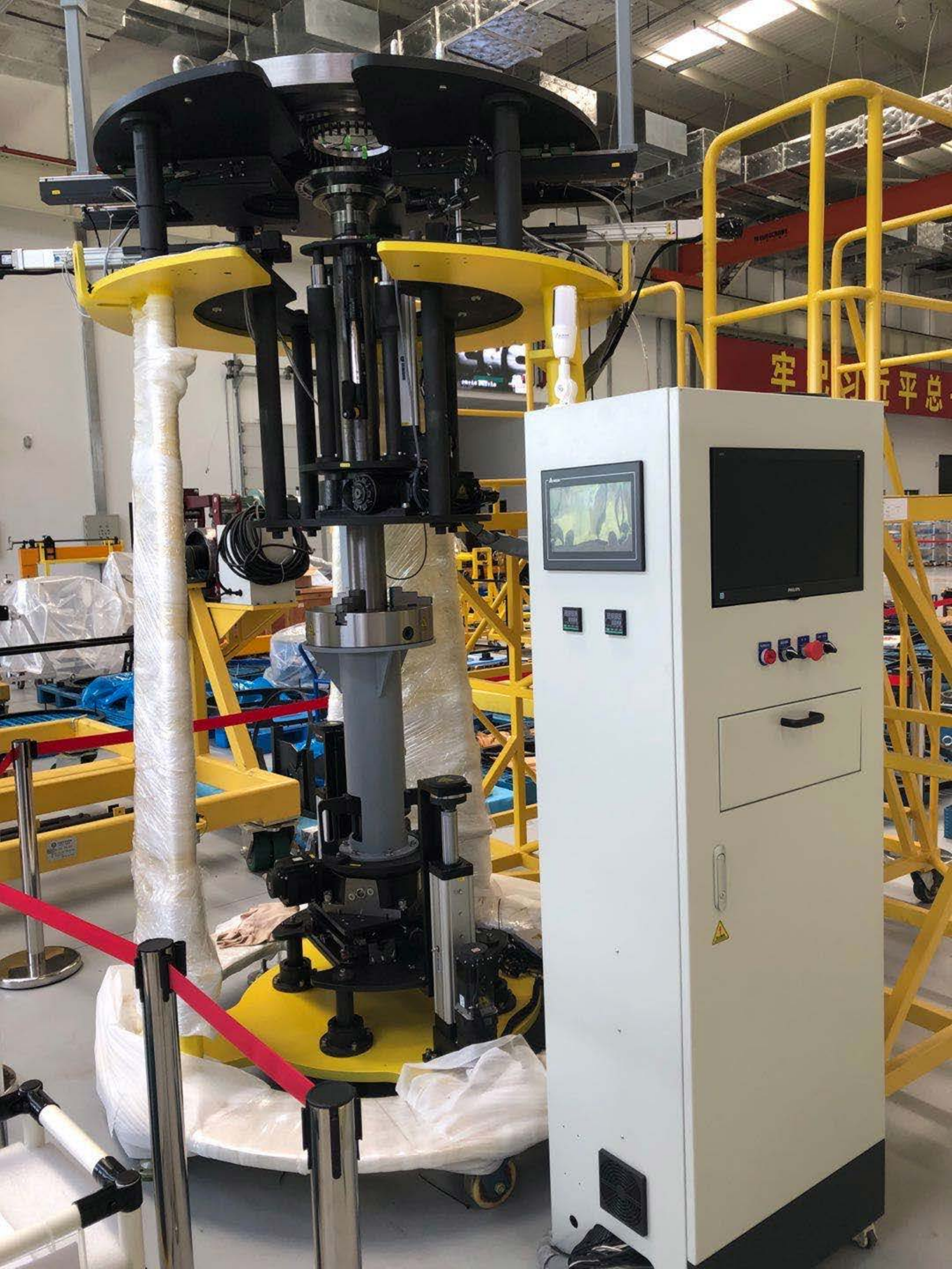}
\caption{Automatic assembly test bed for low pressure turbine shaft}
\label{fig.13}
\end{figure}
According to the functional requirements of automatic assembly of aero-engine low-pressure turbine shaft, the functional modular design is carried out, and the structural design of each module is carried out in a narrow space
\begin{figure}[ht]
\centering
\includegraphics[scale=0.5]{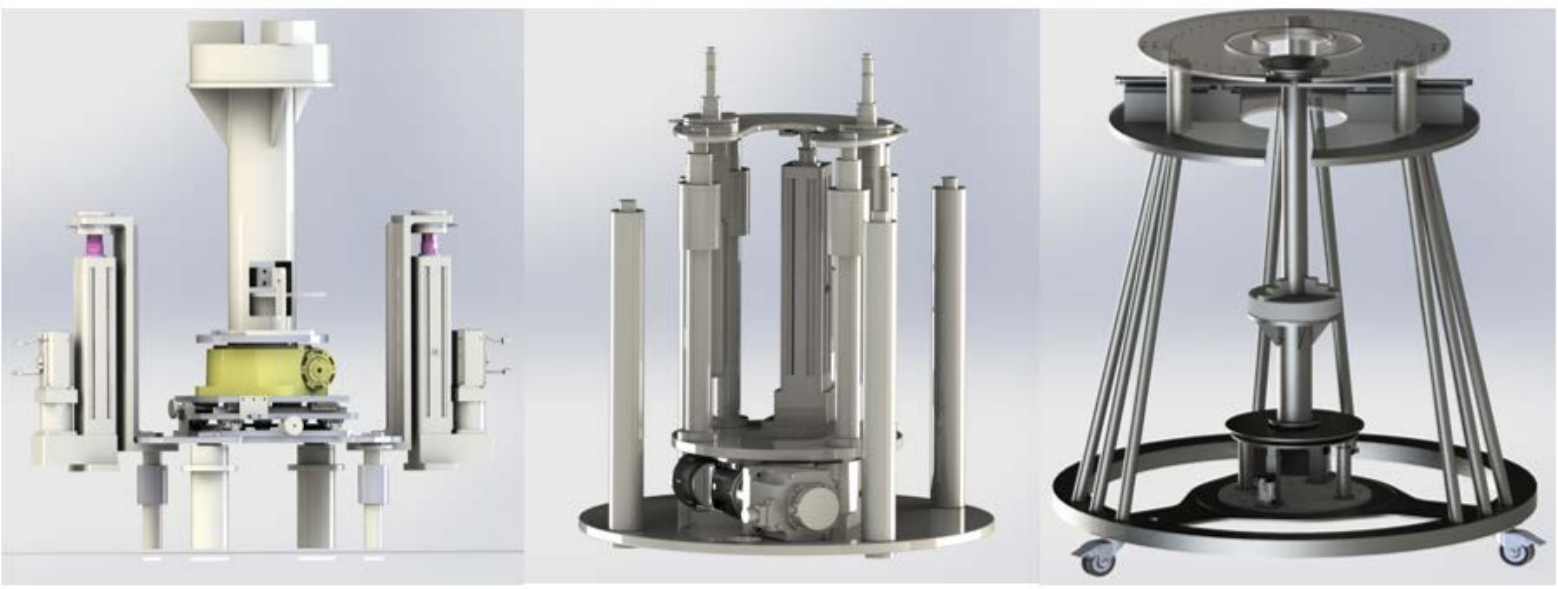}
\caption{Equipment module composition}
\label{fig.10}
\end{figure}

Module 1：four degree of freedom attitude adjustment mechanism，Composition: bilinear module, turbine shaft sleeve, clamping chuck, high precision rotary table, jacking motor，Function: the turbine shaft can be clamped and positioned, which can provide four degrees of freedom attitude adjustment of the turbine shaft; the turbine shaft sleeve and clamping chuck can clamp and center the turbine shaft, the jack up motor drive assembly surface fits, and the precision rotary table drives the turbine shaft hole position alignment

Module 2，Tighten the module；Composition: automatic tightening gun, high precision rotary table, motor, support column, force sensor
Function: two degree of freedom motion mechanism drives the tightening gun to automatically tighten the nut, and the force sensor at the installation position of the tightening gun monitors the contact force in the process of cap recognition

Module 3，Vertical stand and analog parts: three-layer platform, lifting ring, anti torsion support column, universal wheel, etc. functions: the upper platform carries the low-pressure turbine rotor stator, the middle bench is equipped with three-dimensional camera and temperature sensor, and the lower table is installed with attitude adjustment mechanism, and the overall bench is movable

At the assembly site, the visual measurement experiment was carried out with the engine simulator, and the visual measurement results were compared with the laser tracker measurement results
\begin{figure}[ht]
\centering
\includegraphics[scale=0.4]{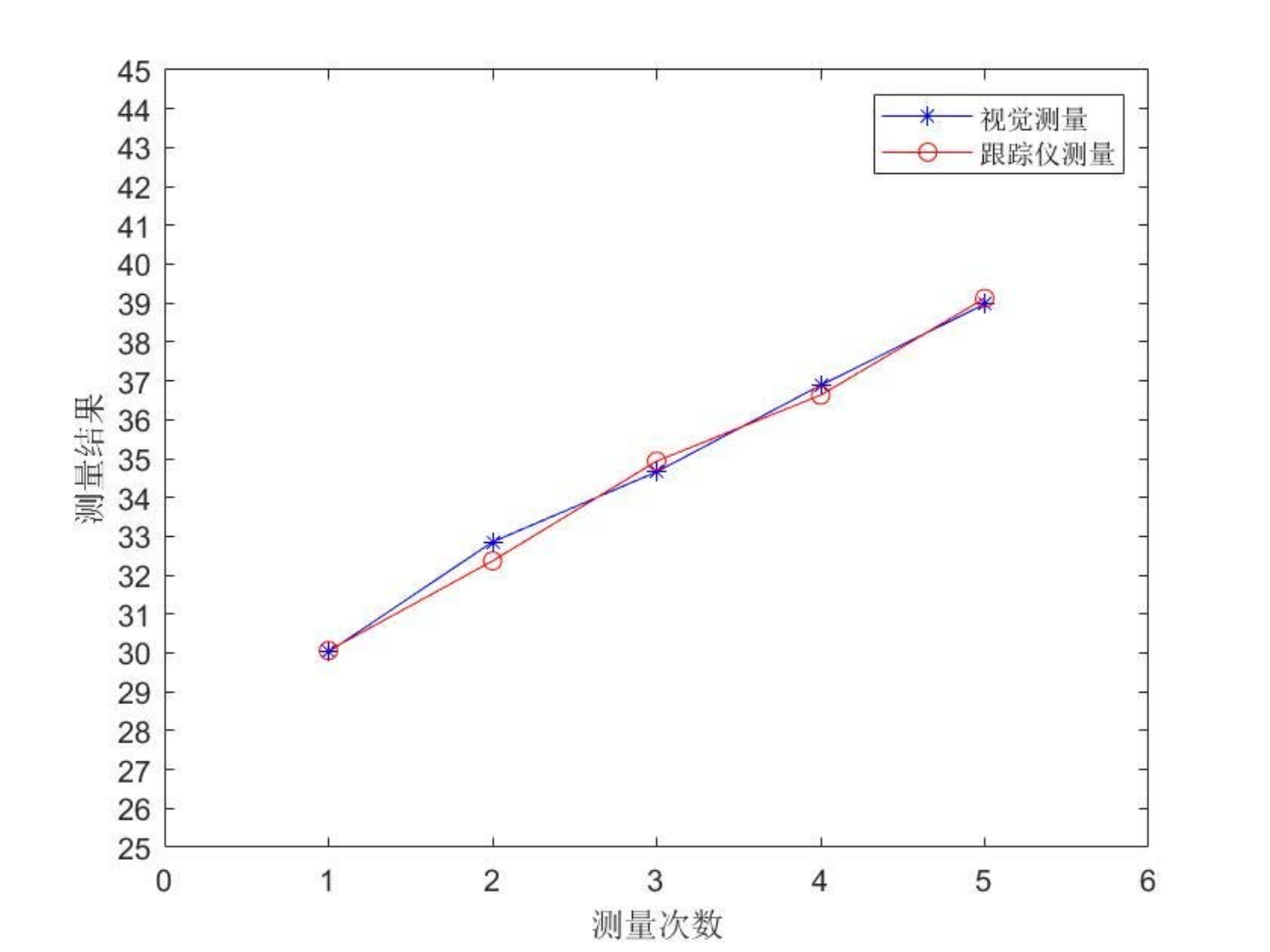}
\caption{Visual measurement results}
\label{fig.11}
\end{figure}
When the installation distance t is within the range of 30-40cm, the measurement accuracy is less than 0.05mm, and the measurement accuracy of hole rotation angle is less than 0.07°，in the range of 0-10 °
\begin{figure}[ht]
\centering
\includegraphics[scale=0.4]{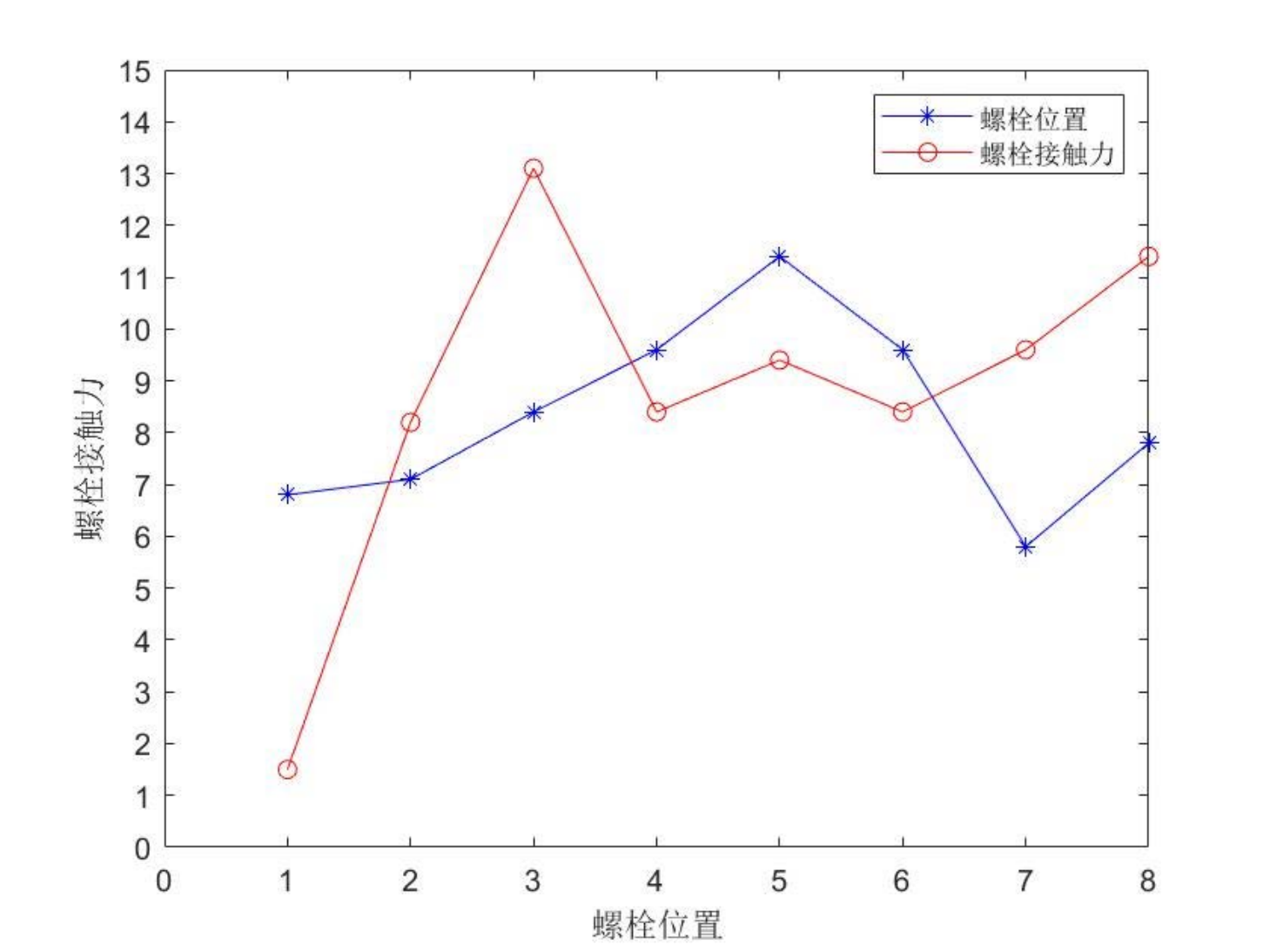}
\caption{Force perception in assembly process}
\label{fig.12}
\end{figure}
According to the process requirements, the evaluation index of successful assembly and docking is that all 36 bolts are inserted into the turbine shaft mounting hole, and the axial impact force of the bolts on the turbine shaft is less than 10N. In the last 30 assembly and docking experiments, all bolts are inserted into the installation hole. The maximum axial force of each bolt is detected by using the pressure sensor. So far, the assembly and docking capacity of the system is estimated the power is ${\rm{86}}{\rm{.7\% }}$.

\section{Conclusion}
The 3D vision system was used to measure and analyze the assembly process of aero-engine low-pressure turbine rotor, which solved the problem of poor light source environment and unable to carry out contact measurement\cite{7}. Because of the high-precision non-contact measurement system, the automatic assembly process of aero-engine low-pressure turbine shaft based on three-dimensional vision is initially established. The key parameters in the measurement algorithm are optimized in the process of engineering application, which improves the assembly power and assembly efficiency of the system\cite{8}. In particular, the specific algorithm flow of 3D thread curve segmentation and extraction modeling for assembly bolt surface is proposed, which solves the measurement problems of hole alignment matching and tightening gun tightening alignment\cite{9}, and establishes the evaluation index of assembly surface matching and hole position alignment matching to optimize measurement, which has universality for solving such engineering problems\cite{10}.

\end{document}